\documentclass{article}

\usepackage{microtype}
\usepackage{graphicx}
\usepackage{subcaption}
\usepackage{booktabs} 

\usepackage{hyperref}

\usepackage{wrapfig}
\usepackage{booktabs}
\usepackage{bm}
\usepackage{multirow}
\usepackage{comment}
\usepackage{subcaption}
\usepackage[normalem]{ulem}
\useunder{\uline}{\ul}{}

\usepackage[ruled,vlined]{algorithm2e}

\usepackage[accepted]{icml2026}

\usepackage{amsmath}
\usepackage{amssymb}
\usepackage{mathtools}
\usepackage{amsthm}

\usepackage[capitalize,noabbrev]{cleveref}

\theoremstyle{plain}

\theoremstyle{definition}

\theoremstyle{remark}

\usepackage[textsize=tiny]{todonotes}

\icmltitlerunning{Building Reliable Long-Form Generation via Hallucination Rejection Sampling}

\begin{document}

\twocolumn[
  \icmltitle{Building Reliable Long-Form Generation via Hallucination Rejection Sampling}



  \icmlsetsymbol{equal}{*}

  \begin{icmlauthorlist}
    \icmlauthor{Lin Li}{oatml}
    \icmlauthor{Georgia Channing}{eng}
    \icmlauthor{Suhaas M Bhat}{oatml}
    \icmlauthor{Gabriel Davis Jones}{dhi}
    \icmlauthor{Yarin Gal}{oatml}
  \end{icmlauthorlist}

  \icmlaffiliation{oatml}{OATML, Department of Computer Science, University of Oxford,
Oxford, UK.}
  \icmlaffiliation{eng}{Department of Engineering, University of Oxford, Oxford, UK.}
  \icmlaffiliation{dhi}{Oxford Digital Health Labs, Nuffield Department of Women’s and
Reproductive Health, University of Oxford, Oxford, UK}

  \icmlcorrespondingauthor{Lin Li}{lin.li@cs.ox.ac.uk}

  \icmlkeywords{Machine Learning, ICML}

  \vskip 0.3in
]



\printAffiliationsAndNotice{}  

\begin{abstract}
Large language models (LLMs) have achieved remarkable progress in open-ended text generation, yet they remain prone to hallucinating incorrect or unsupported content, which undermines their reliability. This issue is exacerbated in long-form generation due to hallucination snowballing, a phenomenon where early errors propagate and compound into subsequent outputs. To address this challenge, we propose a novel inference-time hallucination mitigation framework, named Segment-wise HAllucination Rejection Sampling (SHARS), which uses an arbitrary hallucination detector to identify and reject hallucinated segments during generation and resample until faithful content is produced. By retaining only confident information and building subsequent generations upon it, the framework mitigates hallucination accumulation and enhances factual consistency. To instantiate this framework, we adopt semantic uncertainty as the detector and introduce several vital modifications to address its limitations and better adapt it to long-form text.
Our method enables models to self-correct hallucinations without requiring external resources such as web search or knowledge bases, while remaining compatible with them for future extensions. Empirical evaluations on standardized hallucination benchmarks demonstrate that our method substantially reduces hallucinations in long-form generation while preserving or even improving the informativeness of generation.
Code is available at: \url{https://github.com/TreeLLi/hallucination-rejection-sampling}.



\end{abstract}

\section{Introduction}

Large language models (LLMs) \citep{openai_openai_2025, yang_qwen3_2025, grattafiori2024llama} have markedly expanded the frontiers of artificial intelligence, demonstrating impressive capabilities in open-ended text generation across domains such as question answering \citep{min_factscore_2023, wei_long-form_2024}, code synthesis \citep{jimenezswe}, and scientific research \citep{lu_towards_2026}. 
However, their practical deployment is hindered by a persistent and well-documented challenge: hallucination \citep{ji_survey_2023}. 
Hallucinations arise when models generate content that is factually inaccurate, unsupported, or in conflict with the provided input \citep{bang_hallulens_2025}, often delivered with high fluency. 
This phenomenon undermines the reliability of model output and user trust, and poses risks in high-stakes applications like healthcare and medicine. 

Hallucinations are particularly concerning in open-ended generation, where the extended and unconstrained nature of the outputs makes it especially challenging to validate. In addition, prior studies \citep{zhang2024language, zhao_how_2025, yang_hallucinate_2025} have shown that longer generations tend to amplify hallucination risk, a phenomenon known as \textit{hallucination snowballing}, in which early errors propagate and trigger additional mistakes. This underscores the importance of intervening early in the generation process to interrupt error accumulation and thereby reduce hallucinations.

\begin{figure*}[t]
  \centering
  \begin{subfigure}{0.32\textwidth}
    \includegraphics[width=\linewidth]{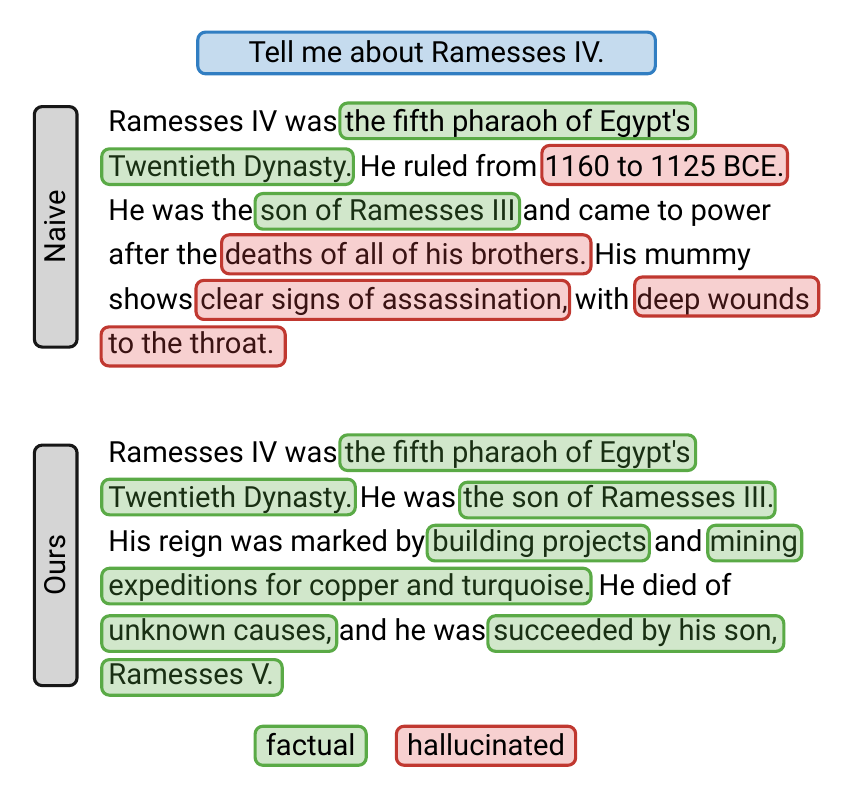}
    \caption{}
    \label{fig:paragraph_compare}
  \end{subfigure}
  \hfill
  \begin{subfigure}{0.32\textwidth}
    \includegraphics[width=\linewidth]{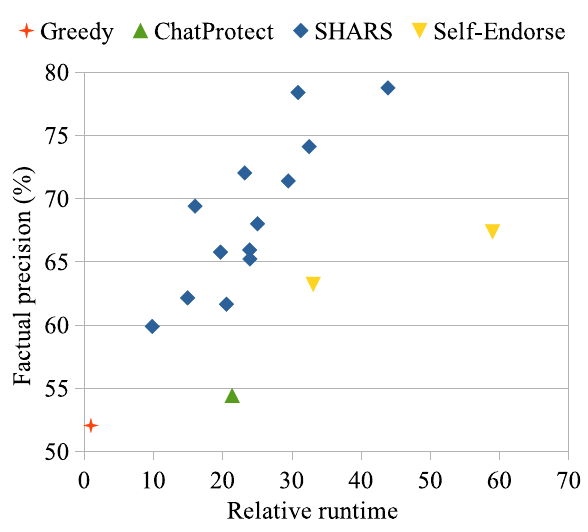}
    \caption{Qwen3-32B}
    \label{fig: efficient scaling qwen}
  \end{subfigure}
  \hfill
  \begin{subfigure}{0.32\textwidth}
    \includegraphics[width=\linewidth]{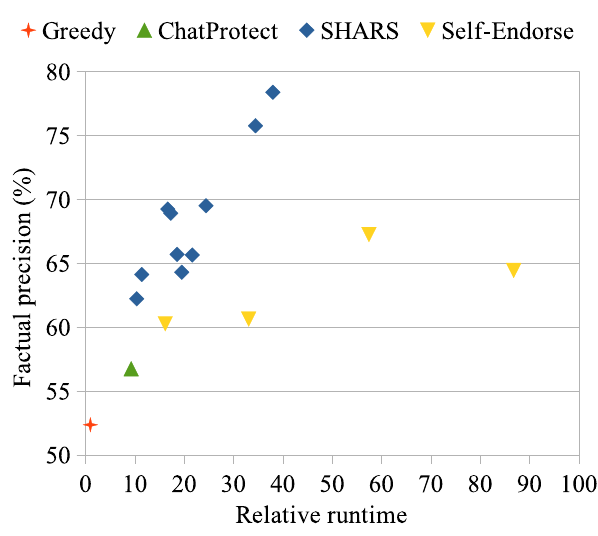}
    \caption{Llama3.1-8B}
    \label{fig: efficient scaling llama}
  \end{subfigure}
  \caption{\textbf{(a)} \textbf{Comparison of biographies generated by Greedy decoding and our method}. Unlike Greedy decoding, our method rejects hallucinated content, preserves factual information, and acquires additional factual content (the last two sentences in the displayed generation) beyond the original information space. 
  \textbf{(b)} and \textbf{(c)} \textbf{Scaling of factual precision with respect to inference-time computation on the FactScore benchmark} \citep{min_factscore_2023}. Inference-time computation is approximated by relative runtime, measured as a factor of the runtime of the corresponding Greedy decoding method for each setup. Full experimental details are provided in \cref{sec: results}.}
  \vspace{-3mm}
\end{figure*}

Separately, a growing body of research \citep{wei_chain--thought_2022,yao_tree_2023,muennighoff_s1_2025,deepseek-ai_deepseek-r1_2025} has investigated the paradigm of inference-time compute scaling, which improves model performance by allocating additional computation at generation time. This paradigm is particularly well-suited for hallucination mitigation in high-stakes domains such as healthcare, scientific discovery, and law, where users are often willing to accept slower responses in exchange for more factual and reliable outputs. Nevertheless, this direction remains underexplored, and to the best of our knowledge, there are no well-established findings on how inference-time scaling affects factuality in open-ended generation.

Inspired by these insights, \textbf{we introduce a general inference-time compute framework, termed Segment-wise HAllucination Rejection Sampling (SHARS), to mitigate hallucinations in open-ended generation}. SHARS leverages an arbitrary detector to identify and reject hallucinated content as it is produced during generation, preserves only factual segments, and builds subsequent outputs upon them (\cref{fig:paragraph_compare}). This design aims to increase the proportion of factual information in the final output while disrupting hallucination snowballing from its early stages. 

To instantiate this framework, \textbf{we further propose a new hallucination detection method, HalluSE, tailored for long-form generation}. HalluSE builds upon the prior semantic entropy approach \citep{farquhar_detecting_2024}, incorporating several refinements to address its limitations and improve detection effectiveness. \textbf{Notably, SHARS is designed to be detector-agnostic, allowing it to integrate with any hallucination detection method} and thereby broadly benefit from future advances in hallucination detection research.

We conduct extensive experiments on diverse long-form factuality benchmarks, including FactualBio \citep{farquhar_detecting_2024}, FactScore \citep{min_factscore_2023}, and LongFact \citep{wei_long-form_2024}, to evaluate our methods. Empirical results show that HalluSE significantly improves hallucination detection accuracy over prior approaches in long-form generation. SHARS further proves effective in mitigating hallucinations in open-ended generation while preserving, and in some cases enhancing, output informativeness. Importantly, \textbf{SHARS exhibits a promising scaling property: when appropriately configured, factuality continues to improve as additional inference-time computation is allocated within a certain range} (\cref{fig: efficient scaling qwen,fig: efficient scaling llama}). For instance, SHARS improves factual precision by about 26\% for evaluated models on the FactScore benchmark.


\section{Related Works}

\textbf{Hallucination mitigation}.
To mitigate hallucinations, \citet{tian_fine-tuning_2024} fine-tune models using preference data generated from a retrieval-enabled judge and direct preference optimization (DPO), enabling the model to prefer factual responses. \citet{huang_factalign_2024} propose FactAlign, which assigns sentence-level factuality rewards to reinforce supported spans in long-form outputs. \citet{gu_mask-dpo_2025} introduce Mask-DPO, which masks non-factual sentences during preference optimization so that updates focus exclusively on factual content.

Inference-time mitigation approaches have also been explored. Integrative decoding \citep{cheng_integrative_2025} aggregates self-consistent continuations by jointly selecting supported tokens. \citet{chuang_dola_2024} propose DoLa, which reweights next-token probabilities by contrasting logits from late and early layers. Retrieval-augmented generation (RAG) \citep{lewis2020retrieval} grounds generation on retrieved passages to replace unsupported spans, while \citet{cai2024forag} improve RAG by introducing outline-guided generation and factuality-aware optimization for web-augmented long-form outputs. More recently, \citet{cheng_think_2025} incorporate tree search–based algorithms to enable explicit slow-thinking generation, mitigating hallucinations during inference.

\textbf{Hallucination detection}.
\citet{farquhar_detecting_2024} introduced hallucination detection via semantic entropy, which estimates uncertainty in the space of meanings by clustering diverse model samples and measuring entropy over the induced semantics. They benchmarked this method against two alternatives: Self-Check \citep{manakul_selfcheckgpt_2023}, where the model verifies its own assertions, and P(True), which measures the probability that the model predicts the token ``True'' when few-shot prompted to compare a main answer with alternatives. Another training- and retrieval-free approach by \citet{mundler_self-contradictory_2024} detects hallucinations by eliciting multiple responses and identifying contradictions or inconsistencies. Other methods train lightweight probes. For instance, \citet{kossen_semantic_2024} trained probes to approximate semantic entropy from hidden states of a single generation, while \citet{obeso_real-time_2025} trained probes on web-search-grounded, entity-level labels to detect hallucinations in real time. Alternatively, \citet{min_factscore_2023}, \citet{wei_long-form_2024}, and \citet{zhao_how_2025} detect hallucinations by decomposing generated text into atomic facts and checking them against trusted external sources.

\section{SHARS: Segment-wise HAllucination Rejection Sampling}
\label{sec: shars}



\textbf{Motivation}.
We observe that open-ended questions often admit an effectively infinite range of relevant information that can constitute a valid answer, yet in practice models draw on only a limited subset of this space when generating responses. Intuitively, if hallucinated content in the initial generation can be filtered out and the model is guided to explore the remaining information space for truthful content to fill these gaps, the resulting generation can be free of hallucinations. Moreover, by dynamically grounding generation on truthful information, this process could potentially disrupt the error compounding caused by earlier mistakes and increase the likelihood of sampling factual content.

\subsection{Segment-wise Hallucination Rejection Sampling}
\label{sec: shars framework}

Following this motivation, we propose our general inference-time compute framework, SHARS, which leverages an arbitrary detector to identify and reject hallucinated content during generation. SHARS partitions the generation into multiple segments and applies hallucination rejection sampling sequentially as each sentence is produced. For a given sentence, hallucination rejection sampling invokes a hallucination detector to assess its factuality. Based on the detection outcome, the sentence is either (i) discarded if it contains no factual information, (ii) rewritten to remove hallucinated content if it mixes factual and hallucinated information, or (iii) retained if it is entirely factual, with no hallucinations detected. Generation terminates when one of the following occurs: (1) an end-of-sequence (EOS) token is sampled; (2) the maximum new-token budget is reached; or (3) fully hallucinated sentences are sampled in $N$ consecutive attempts. The full procedure is summarized in \cref{algo: shars}.

Our method differs from conventional rejection sampling, also known as the best-of-N sampling, for inference-time scaling in three key aspects. First, rejection sampling is performed in a segment-wise and dynamic manner rather than applied once to the entire generation. Second, we sample one candidate sentence at a time and resample only when the current sentence is rejected, instead of generating multiple candidates simultaneously. Third, in cases where a sentence contains both factual and hallucinated information, we rewrite it to remove hallucinations rather than discarding it entirely. The latter two strategies improve efficiency and make it more practical for inference-time deployment.


\begin{algorithm}[tbp]
\DontPrintSemicolon
\KwData{User query $q$}
\KwResult{Verified response to the user query}
verified\_text, hallued\_text $\gets$ ``'', ``''\;

\While{not End\_Of\_Sequence}{
    sent $\gets$ next\_sent($M$, $q$, verified\_text, hallued\_text)\;
    
    verified\_facts, hallued\_facts $\gets$ detect\_hallu($M$, sent, verified\_text)\;
    
    \If{len(verified\_facts) $= 0$}{
        hallued\_text $\gets$ hallued\_text + sent\;
    }
    \Else{
        \If{len(hallued\_facts) $\neq 0$}{
        sent $\gets$ rewrite\_sent($M$, $q$, verified\_facts)\;
    }
    verified\_text $\gets$ verified\_text + sent\;
    
    hallued\_text $\gets$ ``''\;
    }
    \tcp{\small break if no verified\_facts for $N$ times in a row}
}
\Return{verified\_text}
\caption{Pseudocode of SHARS.}
\label{algo: shars}
\end{algorithm}

\textbf{Hallucination detector}.
\textbf{SHARS is designed to operate with any detector by treating the hallucination detector as a black box.} 
In this work, we instantiate SHARS with our entropy-based HalluSE detector (technical details are described later in \cref{sec: method halluse}). We adopt HalluSE because (1) it does not require training a new probe model to detect hallucinations like \citet{obeso_real-time_2025}, and because (2) it does not rely on external tools or reference knowledge sources. These properties allow seamless integration into SHARS and enable zero-shot application across new domains.

We acknowledge that hallucination detection is an active research area, and many alternative methods \citep{aichberger_rethinking_2026, duan_shifting_2024, manakul_selfcheckgpt_2023, chen_inside_2024} may be suitable for integration into our framework. We leave this exploration for future work, as our current choice already achieves substantial reductions in hallucinations and significant improvements in factual precision compared with existing state-of-the-art mitigation methods, as shown in \cref{sec: results hallucination mitigation}.

HalluSE estimates the uncertainty of the knowledge probed by generated questions and uses it as a proxy for the uncertainty of the corresponding fact. For example, consider the fact to be verified: \texttt{Alan Turing is an athlete}. The relevant knowledge in this case is Alan Turing’s profession. If the model is uncertain about this knowledge, it suggests that not only the underlying fact is likely hallucinated, but also that alternative sampled facts for the profession are likely hallucinated. While this improves the efficiency of hallucination detection, it also raises a challenge for sentence sampling: how do we generate a new sentence with knowledge distinct from that in the hallucination?


\textbf{Sentence sampling}.
To address the above challenge, we explore two strategies, termed Temperatures and Following. The Temperatures strategy gradually increases the decoding temperature for sampling a new sentence as the number of consecutive hallucinated sentences grows. In other words, the longer the model is stuck at a given point in generation, the more randomness is introduced to encourage exploration of alternative continuations. This approach leverages the model’s inherent stochasticity to produce diverse sentences, but it can be less efficient as it does not explicitly incorporate information from the identified hallucinated sentences.

In contrast, the Following strategy temporarily retains the identified hallucinated sentences in the generation and samples the next sentence by continuing the generation process, as illustrated in \cref{algo: next sentence}. This leverages the model’s inherent content planning ability to reduce the likelihood of repeatedly generating content about the same knowledge. For example, a model will typically avoid generating a second birthday for an individual once one has already been stated. However, this approach risks allowing hallucinations to influence subsequent generation. To mitigate this effect, we clear the pool of hallucinated sentences whenever new factual information is identified and retained, preventing the pool from becoming excessively large, as shown in \cref{algo: shars}. Furthermore, hallucinated sentences are used solely for sentence sampling and are not passed to HalluSE as context for computing semantic entropy, ensuring that existing hallucinations do not affect the identification of hallucinations in newly sampled sentences. The Following strategy is ultimately adopted due to its superior empirical performance, as discussed in \cref{sec: results ablation study}.


\begin{algorithm}[tbp]
\DontPrintSemicolon
\KwData{$q$, verified\_text, hallued\_text}
\KwResult{A new sentence}
text\_sofar $\gets$ ``''\;
input $\gets$ $q$ + verified\_text + hallued\_text\;
\While{True}{
    token $\gets$ next\_token($M$, input)\;
    
    text $\gets$ decode(token)\;
    
    text\_sofar $\gets$ text\_sofar + text\;
    
    sents $\gets$ split\_sents(text\_sofar)\;
    
    \If{len(sents) $\geq$ 2}{
        sent $\gets$ sents[0]\;
        
        break\;
    }
    input $\gets$ input + text\;
}
\Return{sent}
\caption{Pseudocode of next sentence sampling.}
\label{algo: next sentence}

\end{algorithm}

\textbf{Sentence rewriting}.
We employ an LLM to rewrite the sentence to remove its hallucinated content while preserving factual information. Specifically, we provide the LLM with a list of factual claims identified by HalluSE and prompt it to generate a sentence comprising those claims, rather than supplying the original sentence along with hallucinated claims and asking it to remove them. Empirically, we find that the former approach performs better with small- to medium-scale models such as Qwen3-32B, Llama3.1-8B, and even Qwen3-4B-Instruct. We hypothesize that this advantage arises because LLMs are more effective when guided by positive examples than by negative ones.

The rewriting LLM can be any model with sufficient instruction-following capability to perform the task. In this work, we use the same model as the main generation model. The rewriting prompts are provided in \cref{sec: prompts}.

\subsection{Abstention Mechanism}
Our third termination condition leads to a novel dynamic abstention mechanism based on the model’s parametric knowledge and internal confidence. Assuming sufficient diversity in sentence sampling, our method abstains after generating $N$ fully hallucinated sentences covering different aspects of the user query in a row. This abstention may occur either at the outset or midway through a generation, with the latter case allowing the model to first produce information it is confident is factual.

\section{HalluSE: Detecting Hallucinations in Long-Form Generation}

This section introduces HalluSE, our uncertainty-based hallucination detection method for long-form text generation. HalluSE builds on the prior semantic entropy approach for long-form generation \citep{farquhar_detecting_2024}, while addressing several of its key limitations.

\begin{figure*}
     \centering
     \includegraphics[width=0.99\linewidth]{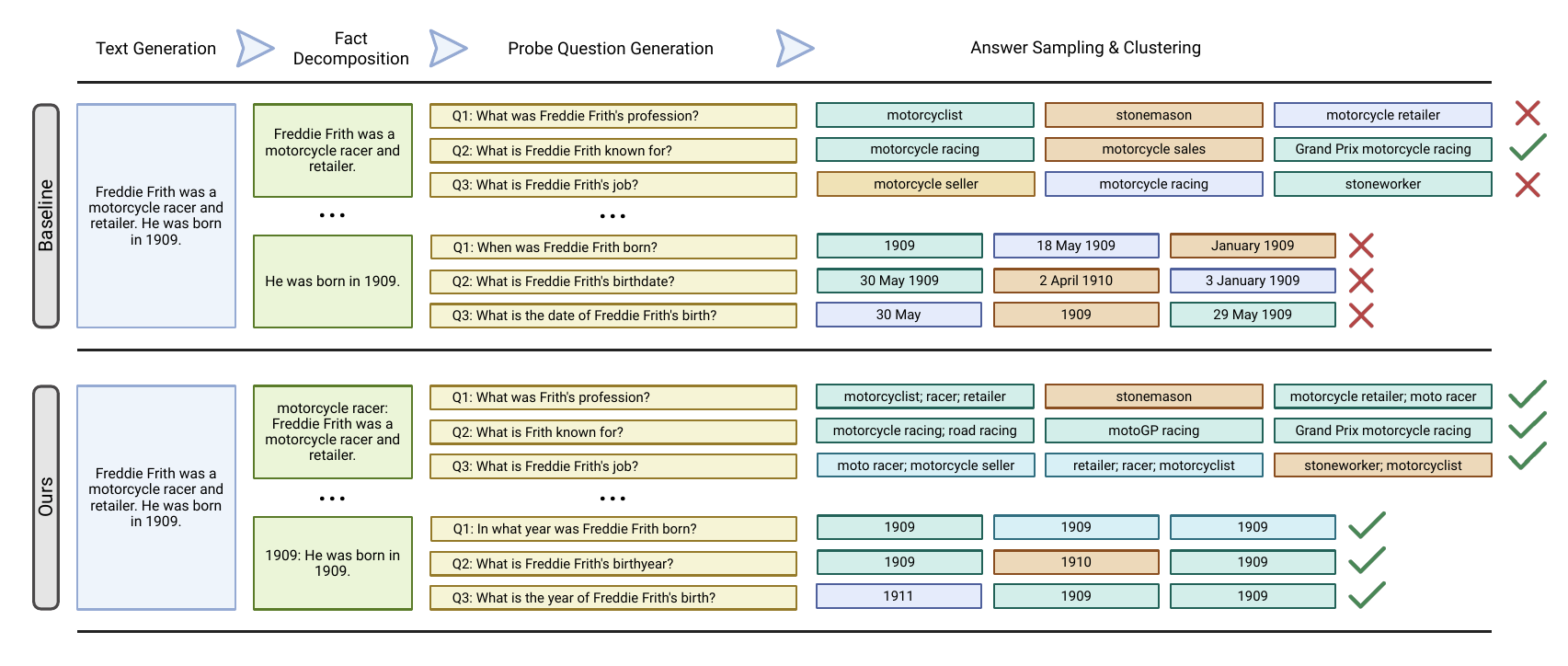}
     \caption{Illustration of naive long-form semantic entropy method and our proposed HalluSE. Different colors under `Answer Sampling \& Response' denote distinct semantic clusters of generated responses. A green check indicates low semantic entropy (high agreement, reliable answers), while a red cross marks high semantic uncertainty (likely hallucinated content).}
     \label{fig: longform hallucination detection pipeplines}
     \vspace{-2mm}
 \end{figure*}

\subsection{Background: Semantic Entropy}
Semantic entropy \citep{farquhar_detecting_2024} is an uncertainty measure that captures the variability of a model’s predictions in the semantic space rather than the token space. 
Instead of only considering surface-level probability distributions over tokens, semantic entropy groups candidate generations into meaning-equivalent clusters and measures the entropy across these clusters. 
Given a set of candidate generations sampled from the model, each generation is mapped into a semantic cluster $C_i$. The probability of a cluster is defined as the sum of token-level probabilities of all generations assigned to it: $p(C_i) = \sum_{y \in C_i} p(y)$, where $p(y)$ is the model probability of generation $y$. 
The semantic entropy is then computed as the entropy over cluster probabilities: $H_s = - \sum_{i} p(C_i) \log p(C_i)$. The full technical details can be found in \citet{farquhar_detecting_2024}.

Low semantic entropy indicates semantic agreement among candidate generations, while high semantic entropy reflects semantic disagreement and is often associated with hallucinations. This property makes semantic entropy a natural signal for hallucination detection: when the model is confident and semantically consistent, the likelihood of hallucination is lower, whereas high semantic entropy often correlates with unsupported or erroneous content.

\subsection{Naive Long-Form Semantic Entropy}
The semantic entropy method described above assumes that candidate answers are short-form. To extend it, \citet{farquhar_detecting_2024} proposed a naive approach for applying semantic entropy to long-form generation. As shown in \cref{fig: longform hallucination detection pipeplines}, the generation is first decomposed into a set of fact claims. For each fact claim, several probe questions with expected answers are generated to query the fact, and the short-form semantic entropy method is then applied to each question. This procedure effectively reduces long-form hallucination detection to a series of short-form detection tasks.

The naive long-form semantic entropy method faces two main limitations as illustrated in \cref{fig: longform hallucination detection pipeplines}. First, it decomposes a generation into fact claims without distinguishing which entity within each claim should be validated. This ambiguity can cause the wrong entity to be probed downstream. For example, given the query \texttt{Tell me about Alan Turing} and the generation \texttt{Alan Turing is a computer scientist}, the entity of interest is clearly \texttt{computer scientist} rather than \texttt{Alan Turing}. However, the prior method may incorrectly generate probe questions such as \texttt{Who is a computer scientist?}.

Second, the naive approach assumes that each probe question has only a single valid answer, so any uncertainty in sampled answers is attributed solely to the model. In practice, however, probe questions can admit multiple valid answers. For example, in biographies, a prominent individual may hold multiple professions. A probe question such as \texttt{What is XX’s profession?} may thus have several correct answers. Even if the model consistently samples correct but different professions, the resulting semantic entropy remains high, incorrectly flagging the fact as hallucinated.

\subsection{HalluSE}
\label{sec: method halluse}

HalluSE addresses the limitations of the naive long-form semantic entropy method through three key refinements as illustrated in \cref{fig: longform hallucination detection pipeplines}. First, it decomposes each generation into pairs of entities and fact claims. Second, it improves the prompting strategy with clearer instructions, structured formatting, and few-shot examples. In particular, HalluSE guides the LLM to generate probe questions with unambiguous expected answers, thereby reducing unnecessary cases of multiple valid answers. Third, it explicitly instructs the LLM to provide all valid answers, when applicable, in each sampling step. The complete HalluSE pipeline is as follows:

\begin{enumerate}
    \item \textbf{Fact Decomposition}: given a generation, HalluSE decomposes it into a set of facts, where each fact consists of an entity and a claim describing a piece of atomic information about that entity from the model response.

    \item \textbf{Question Generation}: for each fact, HalluSE generates $Q$ probe questions in which the entity and claim serves as the expected short-form and long-form answer, respectively.

    \item \textbf{Answer Sampling}: for each probe question, it produces $A$ answers conditioned on the preceding context $c$, i.e., the text appearing before the fact in the response.

    \item \textbf{Semantic Entropy Computation}: semantic entropy is computed from the sampled answers per question and averaged across the $Q$ questions, yielding the semantic entropy of the fact.

    \item \textbf{Hallucination Identification}: A fact is classified as hallucinated if its semantic entropy exceeds a predefined threshold $\theta$; otherwise, it is deemed factual. 
\end{enumerate}

The full procedure is summarized in \cref{algo: HalluSE}. Fact Decomposition, Question Generation, and Answer Sampling are implemented by prompting a pretrained instruction-following LLM, denoted as $M$, with the specific prompts detailed in \cref{sec: prompts}. 
Semantic Entropy Computation is implemented with its discrete formulation \citep{farquhar_detecting_2024}.
The LLMs for Fact Decomposition and Question Generation can be arbitrary, while the LLM for Answer Sampling should match the model used to produce the given response. 
In this work, we employ the same model for all components, including response generation.

\begin{algorithm}[tbp]
\DontPrintSemicolon
\KwData{text, $c$, $M$, $Q$, $A$, $\theta$}
\KwResult{verified facts, hallucinated facts}
verified\_facts, hallued\_facts $\gets$ [ ], [ ]\;

facts $\gets$ decompose\_facts($M$, text)\;

\For{(entity, claim) in facts}{
    questions $\gets$ gen\_questions($M$, $Q$, entity, claim)\;
    
    $H_s$ $\gets$ [ ]\;
    
    \For{question in questions}{
        $H_s$ $\gets$ $H_s$ $\cup$ semantic\_entropy($M$, $A$, $c$, question)\;
    }
    $H_s$ $\gets$ mean($H_s$)\;
    \If{$H_s < \theta$}{
        verified\_facts $\gets$ verified\_facts $\cup$ (entity, claim)\;
    }
    \Else{
        hallued\_facts $\gets$ hallued\_facts $\cup$ (entity, claim)\;
    }
}
\Return{verified\_facts, hallued\_facts}
\caption{Pseudocode of HalluSE.}
\label{algo: HalluSE}
\end{algorithm}

\begin{table}[h]
\centering
\caption{Performance of baselines and our methods on the FactScore benchmark without constraints on response length. The best score for each metric is highlighted in \textbf{bold}.}
\label{tab: factscore}
\begin{tabular}{@{}lcccc@{}}
\toprule
\multicolumn{1}{c}{Method} & Resp. (\%)     & Unsup.       & Supp.         & FPrec. (\%)   \\ \midrule
\multicolumn{5}{c}{Qwen3-4B}                                                              \\ \midrule
Greedy                      & 91.2          & 11.2         & 11.2          & 50.0          \\
DoLa                       & \textbf{94.5} & 11.1         & 11.9          & 51.8          \\
ChatProtect                & 91.8          & 9.7          & 10.8          & 52.6          \\
Self-Endorse               & 92.3          & 6.6          & 9.2           & 58.2          \\ \midrule
Ours-Resp                  & 92.9          & 8.8          & 14.9          & 63.0          \\
Ours-Info                  & 89.0          & 8.2          & 16.3          & 66.6          \\
Ours-Prec                  & 69.8          & \textbf{5.7} & \textbf{16.2} & \textbf{74.0} \\ \midrule
\multicolumn{5}{c}{Llama3.1-8B}                                                           \\ \midrule
Greedy                      & \textbf{99.5} & 5.7          & \textbf{6.7}  & 53.7          \\
DoLa                       & \textbf{99.5} & 5.7          & \textbf{6.7}  & 53.8          \\
ID                         & 98.3          & 5.0          & 5.7           & 53.5          \\
ChatProtect                & 97.3          & 5.0          & 6.5           & 56.7          \\
Self-Endorse               & 96.7          & 4.1          & 6.3           & 60.6          \\ \midrule
Ours-Resp                  & \textbf{99.5} & 3.2          & 5.7           & 64.1          \\
Ours-Info                  & 88.5          & 1.9          & 5.9           & 75.6          \\
Ours-Prec                  & 78.6          & \textbf{1.4} & 5.0           & \textbf{78.4} \\ \midrule
\multicolumn{5}{c}{Qwen3-32B}                                                             \\ \midrule
Greedy                      & \textbf{99.5} & 8.8          & 9.7           & 52.4          \\
DoLa                       & 95.6          & 9.3          & 8.2           & 53.1          \\
ChatProtect                & 98.9          & 8.1          & 6.8           & 54.4          \\
Self-Endorse               & 91.8          & 4.9          & 8.4           & 63.2          \\ \midrule
Ours-NE                  & 97.3 & 4.2 & 9.9 & 70.1 \\
Ours-Resp                  & 97.8          & 5.7          & 11            & 65.7          \\
Ours-Info                  & 92.9          & 4.2          & \textbf{11.7} & 73.5          \\
Ours-Prec                  & 82.4          & \textbf{3.1} & 11.1          & \textbf{78.4} \\ \bottomrule
\end{tabular}
\vspace{-7mm}
\end{table}

\section{Results}
\label{sec: results}
\textbf{Experiment setup.} We mainly evaluate our method on the FactScore benchmarks using Qwen3-4B, Qwen3-32B \citep{yang_qwen3_2025} and Llama3.1-8B-Instruct \citep{grattafiori2024llama}.
Qwen3-4B/32B follows officially recommended decoding settings with temperature 0.7, top-$k$ 20, and top-$p$ 0.8, while Llama3.1-8B-Instruct uses temperature 0.7, top-$k$ 50, and top-$p$ 0.9. For baselines, we use Greedy decoding, DoLa \citep{chuang_dola_2024}, ID \citep{cheng_integrative_2025}, ChatProtect \citep{mundler_self-contradictory_2024}, and Self-Endorse \citep{wang_improving_2024}. The ID results for Qwen3 models are omitted because the authors’ released code is incompatible with Qwen3. The full experiment setup and the configuration of our methods are given in \cref{app: configuration}.


Factual precision (``FPrec.'') is defined as the proportion of supported claims (``Supp.'') relative to the total number of claims (``Supp.'' + ``Unsup.''). 
Response rate (``Resp.'') denotes the proportion of queries answered without refusal. 
Factual precision and the number of fact claims are computed with generations that answer without refusal. 

For each model, results of our method are reported under three hyperparameter settings: Ours-Resp maximizing the response rate, Ours-Info maximizing the number of supported claims, and Ours-Prec maximizing factual precision. 
In addition, we also replace semantic entropy (SE) with token-level naive entropy (NE) to measure uncertainty, named Ours-NE, while keeping the rest of the pipeline unchanged.

\subsection{Reduced Hallucinations \& Increased Information}
\label{sec: results hallucination mitigation}

\begin{table}[]
\centering
\caption{Performance of our methods for Qwen3-32B on the FactScore benchmark with a 200-word response length constraint. Models are prompted to generate around 200 words, which exceeds the average length produced without such constraint.}
\label{tab: factscore 200words}
\begin{tabular}{@{}lcccc@{}}
\toprule
\multicolumn{1}{c}{Method} & Resp. (\%)     & Unsup.       & Supp.         & FPrec. (\%)   \\ \midrule
Greedy                      & \textbf{98.9} & 16.2         & 22.4          & 58.0          \\
DoLa                       & 97.8          & 16.9         & 22.3          & 56.9          \\
ChatProtect                & 97.8          & 14.7         & 21.3          & 59.2          \\ \midrule
Ours-Info                  & 98.4          & 11.8         & \textbf{29.1} & 71.0          \\
Ours-Prec                  & 84.6          & \textbf{6.7} & 23.6          & \textbf{77.9} \\ \bottomrule
\end{tabular}
\vspace{-2mm}
\end{table}

\textbf{Reduced hallucination rate}. As shown in \cref{tab: factscore,tab: factscore 200words}, our method substantially reduces hallucination rates across different models and generation lengths. It consistently improves factual precision over the Greedy baseline by approximately 20–26\% and significantly decreases the number of unsupported fact claims that are hallucinated by the model.

Although NE underperforms SE (70.1\% vs. 78.4\% FPrec.), it still surpasses the strongest baseline, Self-Endorse (63.2\% FPrec), indicating that SHARS’ effectiveness stems from both the overall framework and the improved detector.

\textbf{Increased factual information}. In addition, \cref{tab: factscore,tab: factscore 200words} show that our method increases the number of supported fact claims across all setups with Qwen3-32B, indicating that the generated responses contain more factual information and are thus more helpful. For Llama3.1-8B-Instruct, our method slightly reduces supported fact claims, but this is minor compared to the substantial reduction in hallucination.

\textbf{Abstention}. We observe that our method achieves the highest factual precision and the largest number of supported facts, albeit with a lower response rate. This indicates that the method effectively identifies user queries for which the underlying model has limited knowledge and abstains from answering. To further validate its effectiveness in mitigating hallucinations independent of additional abstention, we report results under a matched response rate with the baseline, denoted Ours-Resp in \cref{tab: factscore} and Ours-Info in \cref{tab: factscore 200words}. Even under this setting, our method substantially improves factual precision compared to the baseline.

\begin{figure}[tbp]
    \centering
    \includegraphics[width=.8\linewidth, trim=20 10 0 0]{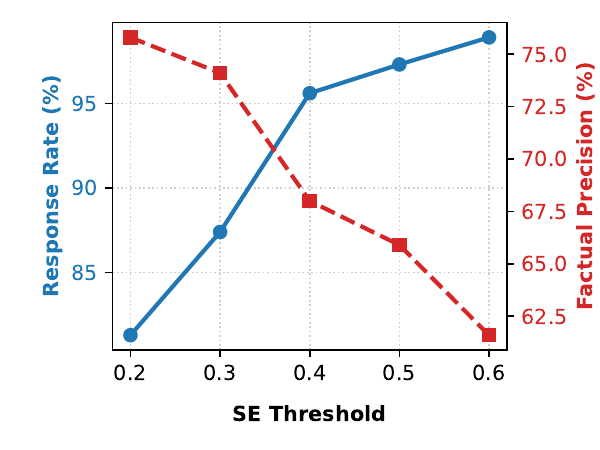}
    \caption{The change of response rate and factual precision with the semantic entropy threshold $\theta$.}
    \label{fig: threshold ablation}
    \vspace{-5mm}
\end{figure}

We further examine factual precision (risk) as a function of response rate (coverage) by varying the uncertainty threshold for detection in \cref{fig: threshold ablation}. We observe a clear and smooth trade-off. This demonstrates that the gains are not solely due to extreme abstention, but reflect a controllable trade-off where reliability improves consistently as lower-confidence outputs are filtered. Importantly, even in the high-coverage regime (Resp. $>$ 95\%), SHARS still achieves substantial improvements, indicating that gains persist beyond selective non-response. Moreover, since abstention is triggered by repeated failures to generate verified content, this behavior suggests that SHARS naturally abstains on queries where the model lacks reliable knowledge, rather than arbitrarily rejecting responses.

To assess consistency, we repeat the experiments for Qwen3-32B and Llama3.1-8B in \cref{tab: factscore} three times and measure the mean and standard deviation in \cref{tab: std} in \cref{app: standard deviation}, which confirm the robustness and consistency of our results.

\begin{table}[tbp]
\centering
\caption{Performance of combining FactAlign with our method on the FactScore benchmark for the Llama3-8B-Instruct model.}
\label{tab: factalign}
\begin{tabular}{@{}lcccc@{}}
\toprule
\multicolumn{1}{c}{Method} & Resp. (\%)      & Unsup.       & Supp.         & FPrec. (\%)   \\ \midrule
\multicolumn{5}{c}{No Response Length Constraint}                                          \\ \midrule
FactAlign                  & \textbf{100.0} & 4.2          & \textbf{4.7}  & 53.1          \\
+ Ours-Resp                & 98.4           & 1.7          & 3.7           & 69.1          \\
+ Ours-Prec                & 73.6           & \textbf{0.9} & 3.6           & \textbf{80.6} \\ \midrule
\multicolumn{5}{c}{200-word Response Length Constraint}                                    \\ \midrule
FactAlign                  & \textbf{100.0} & 13.6         & \textbf{16.9} & 55.4          \\
+ Ours-Resp                & \textbf{100.0} & 8.6          & 14.6          & 62.9          \\
+ Ours-Prec                & 78.6           & \textbf{4.0} & 15.3          & \textbf{79.1} \\ \bottomrule
\end{tabular}
\vspace{-3mm}
\end{table}

\subsection{Generalization to Small-Scale Models}
To further validate effectiveness on smaller models, we additionally evaluated our method using Qwen3-4B. It achieved comparable gains in factual precision (+16–24\%), as shown in Table \ref{tab: factscore}. We note that models with more than 4B parameters already cover a wide range of architectures used in both academic research and real-world applications. These results indicate that, although SHARS involves multiple LLM calls, it does not rely heavily on strong instruction-following capabilities. In other words, the effectiveness of the method does not diminish for smaller or less capable models.

\subsection{Complementary to Training-Time Methods}

The results in Table \ref{tab: factalign} show that our method provides strong complementary benefits to the training-time hallucination mitigation method FactAlign \citep{huang_factalign_2024}. Under both unconstrained and length-constrained settings, adding Ours-Resp consistently reduces unsupported claims and improves factual precision, while Ours-Prec achieves the largest gains—boosting precision from 53.1\% to 80.6\% without length constraints and from 55.4\% to 79.1\% with a 200-word limit. These results indicate that combining FactAlign with SHARS produces a more effective and reliable hallucination-mitigation strategy than FactAlign alone.

\subsection{Efficient Scaling of Factual Precision}

Across both figures in \cref{fig: efficient scaling qwen,fig: efficient scaling llama}, \textbf{SHARS consistently demonstrates a far better accuracy–efficiency trade-off than the baselines}. In the Qwen3-32B plot, SHARS points cluster in the region of relatively low runtime (roughly 10–40×) while achieving higher factual precision (around 60–78\%). In contrast, Self-Endorse requires much higher runtime (about 35–60x) to reach similar or even lower precision, and ChatProtect incurs additional computational cost with only limited improvement over the Greedy baseline. The trend is even more pronounced for Llama3.1-8B: SHARS maintains strong precision (around 68–78\%) at runtimes below 50x, whereas Self-Endorse pushes beyond 60x runtime to achieve comparable accuracy. Taken together, the results show that SHARS delivers higher precision at substantially lower computational cost, making it significantly more efficient than existing methods.

\subsection{Long-Form Hallucination Detection}
This section compares HalluSE against long-form semantic entropy and other closely related detection baselines.

\begin{table}[tbp]
\centering
\caption{Performance of the baseline and our methods on the FactualBio benchmark with Qwen3-32B when detecting both Major and Minor hallucinations.}
\label{tab: factualbio}
\begin{tabular}{@{}lcccc@{}}
\toprule
Method & AUROC & AURAC & Acc@$0.8$ & Acc@$0.9$ \\
\midrule
Self-Check & 57.6 & 69.3 & 73.5 & \textbf{73.5} \\
P(True) & 69.8 & 73.3 & 70.0 & 70.0 \\
Naive SE & 66.2 & 73.1 & 70.5 & 70.5 \\
Ours & \textbf{72.9} & \textbf{77.3} & \textbf{75.4} & 72.8 \\
\bottomrule
\end{tabular}
\vspace{-3mm}
\end{table}

\textbf{Experiment setup.} We evaluate our hallucination detection method \textbf{without SHARS applied} for Qwen3-32B ~\cite{yang_qwen3_2025} on the FactualBio dataset introduced by ~\citet{farquhar_detecting_2024}. FactualBio contains paragraph-length biographies of 21 individuals sampled from the WikiBio dataset~\cite{lebret2016neuraltextgenerationstructured_wikibio}. Each paragraph-length biography in the FactualBio dataset is broken down into individual sentences, which are labeled True, Incorrect-Minor, or Incorrect-Major, depending on the severity of the false claim. For example, the claim that an individual was knighted, though they were not, is considered Incorrect-Major, while a reported birthdate in the wrong month is considered Incorrect-Minor. The same answers generated by GPT-4 were evaluated for different detection methods.

To benchmark our hallucination detection method, we extended the FactualBio dataset to include ``entities``, with respect to which our method evaluates the semantic uncertainty of each claim. The Self-Check baseline, rather than evaluating semantic uncertainty, simply asks the LLM whether the factoid is likely to be true. The P(True) baseline considers the probability that the LLM predicts that the next token is ``True'' when few-shot prompted to compare the original answer with plausible alternatives. 

AURAC summarizes how much accuracy improves when discarding the most uncertain answers. Accuracy@$0.8$/$0.9$ report the model's accuracy after discarding the top 20\% and top 10\% most uncertain responses, respectively.


\textbf{Improved detection accuracy.} In Table~\ref{tab: factualbio}, we observe that our method improves hallucination detection AUROC significantly. AUROC measures how well the method distinguishes correct from incorrect answers across all thresholds.

\subsection{Additional Results on LongFact benchmark}

\begin{table}[]
\centering
\caption{Performance of the baseline and our methods on the LongFact benchmark under different generation length constraints.}
\label{tab: longfact}
\begin{tabular}{@{}lcccc@{}}
\toprule
\multicolumn{1}{c}{Method} & Resp. (\%) & Unsup.       & Supp.         & FPrec. (\%)   \\ \midrule
\multicolumn{5}{c}{No Response Length Constraint}                                     \\ \midrule
Greedy                     & 100.0     & 1.7          & \textbf{23.1} & 93.0          \\
Ours                       & 100.0     & \textbf{1.1} & 21.2          & \textbf{94.6} \\ \midrule
\multicolumn{5}{c}{200-word Response Length Constraint}                               \\ \midrule
Greedy                     & 100.0     & 3.2          & \textbf{43.4} & 93.0          \\
Ours                       & 100.0     & \textbf{2.5} & 41.8          & \textbf{94.4} \\ \bottomrule
\end{tabular}
\vspace{-3.2mm}
\end{table}

In addition to FactScore, we evaluate our method on an alternative long-form factuality benchmark, LongFact \citep{wei_long-form_2024}. As shown in \cref{tab: ablation study}, our method consistently mitigates hallucinations on LongFact, improving factual precision and reducing unsupported fact claims compared to the baseline. Although the improvement margin is smaller than in the FactScore experiments, this is expected since the baseline already achieves very high factual precision. Notably, the 1.4\% precision gain from our method is comparable to the 0.9\% improvement observed when moving from GPT-3.5-Turbo to GPT-4-Turbo, as reported in \citet{wei_long-form_2024}. We emphasize that the reported results are based on a single run without hyperparameter tuning due to the high API cost of evaluation, suggesting that further gains are likely achievable with hyperparameter optimization.



\subsection{Ablation Study}
\label{sec: results ablation study}

This section presents an ablation study on two components of our method: sentence sampling and rewriting. As shown in \cref{tab: ablation study}, both Rewriting and the Following sampling strategy are critical for achieving strong performance. Enabling rewriting substantially boosts the response rate, while the Following strategy increases the number of supported fact claims and, when combined with rewriting, further improves factual precision.

\begin{table}[tbp]
\centering
\caption{Performance of various variants of our method on the FactScore benchmark for Qwen3-32B model. No generation length constraint was applied. All variants were evaluated with the same hyperparameter settings described in \cref{app: configuration}. Relative runtime is reported as a factor with respect to the runtime of the Following-Rewrite variant.}
\label{tab: ablation study}
\resizebox{\linewidth}{!}{%
\begin{tabular}{@{}lcccccc@{}}
\toprule
\multicolumn{1}{c}{Sentence} & \multirow{2}{*}{Rewrite} &  Resp.     & \multirow{2}{*}{Unsup.}          & \multirow{2}{*}{Supp.}           & FPrec.        & Relative      \\
\multicolumn{1}{c}{Sampling} &                          &  (\%)   &   &      & (\%) & Runtime       \\ \midrule
Following                    & Yes                      & 91.8          & 4.8          & 10.7          & 69.4           & \textbf{1.00} \\
Temperature                  & Yes                      & \textbf{95.6} & 4.9          & 9.0           & 64.8           & 1.01          \\
Following                    & No                       & 54.4          & 4.3          & \textbf{12.0} & 73.5           & 1.60          \\
Temperature                  & No                       & 40.1          & \textbf{2.3} & 7.4           & \textbf{76.2}  & 1.55          \\ \bottomrule
\end{tabular}
}
\end{table}

\begin{figure}[tbp]
    \centering
    \includegraphics[width=.8\linewidth, trim=20 10 0 0]{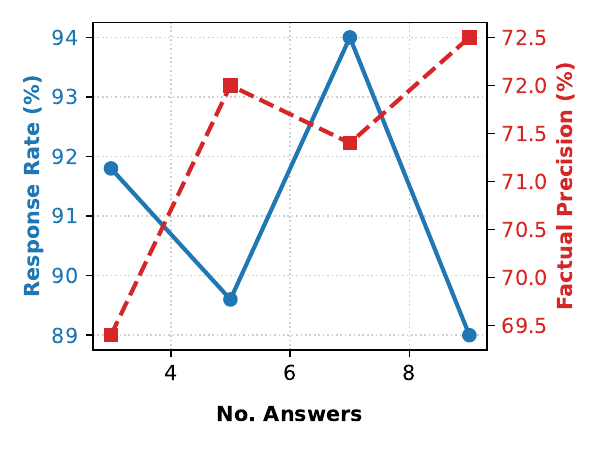}
    \caption{The change of response rate and factual precision with the number of answers, $A$.}
    \label{fig: num answer ablation}
\end{figure}

We additionally report sensitivity to the hyperparameter, the number of sampled answers $A$ in \cref{fig: num answer ablation}, showing that performance is relatively stable across the evaluated range.

\section{Limitations}

We acknowledge two limitations in our method. First, prioritizing factuality may occasionally sacrifice informativeness, though our method empirically maintains a superior balance of these attributes compared to state-of-the-art baselines. Second, our method does not retrieve additional information from external resources, and thus cannot enable the model to answer questions for which it has no faithful knowledge. For example, if a model knows nothing about a person A, rejection and resampling alone will not produce new or accurate information. Nevertheless, significant value remains in the method's ability to drastically reduce hallucinations in scenarios where the model possesses partial knowledge, alleviating the users' effort in auditing the model output.



\section{Conclusion}
In conclusion, this work addresses the critical challenge of hallucinations in open-ended generation by introducing SHARS, a general inference-time compute framework that incrementally rejects hallucinated content and builds subsequent outputs upon verified information. Together with HalluSE, our improved uncertainty-based detection method, SHARS provides an effective and flexible approach to mitigating hallucinations while maintaining or enhancing informativeness. Extensive evaluations across multiple long-form factuality benchmarks demonstrate that our methods significantly advance the state of hallucination detection and mitigation, and importantly, reveal a promising inference-time scaling property of factuality. These findings highlight the potential of inference-time compute as a powerful and practical paradigm for improving the reliability of large language models, especially in high-stakes domains where accuracy and trustworthiness are paramount.

\section*{Acknowledgements}
This work was funded under the Horizon Europe grant 101213369 DVPS. This study was also supported by the UK Medical Research Council, grant number MR/X029689/1.

\section*{Impact Statement}
This paper presents work whose goal is to advance the field of Machine
Learning. There are many potential societal consequences of our work, none
which we feel must be specifically highlighted here.

\bibliography{references,extra_refs}

@article{grattafiori2024llama,
  title={The llama 3 herd of models},
  author={Grattafiori, Aaron and Dubey, Abhimanyu and Jauhri, Abhinav and Pandey, Abhinav and Kadian, Abhishek and Al-Dahle, Ahmad and Letman, Aiesha and Mathur, Akhil and Schelten, Alan and Vaughan, Alex and others},
  journal={arXiv preprint arXiv:2407.21783},
  year={2024}
}

@misc{lebret2016neuraltextgenerationstructured_wikibio,
      title={Neural Text Generation from Structured Data with Application to the Biography Domain}, 
      author={Remi Lebret and David Grangier and Michael Auli},
      year={2016},
      eprint={1603.07771},
      archivePrefix={arXiv},
      primaryClass={cs.CL},
      url={https://arxiv.org/abs/1603.07771}, 
}

@article{lewis2020retrieval,
  title={Retrieval-augmented generation for knowledge-intensive nlp tasks},
  author={Lewis, Patrick and Perez, Ethan and Piktus, Aleksandra and Petroni, Fabio and Karpukhin, Vladimir and Goyal, Naman and K{\"u}ttler, Heinrich and Lewis, Mike and Yih, Wen-tau and Rockt{\"a}schel, Tim and others},
  journal={Advances in neural information processing systems},
  volume={33},
  pages={9459--9474},
  year={2020}
}

@inproceedings{cai2024forag,
  title={Forag: Factuality-optimized retrieval augmented generation for web-enhanced long-form question answering},
  author={Cai, Tianchi and Tan, Zhiwen and Song, Xierui and Sun, Tao and Jiang, Jiyan and Xu, Yunqi and Zhang, Yinger and Gu, Jinjie},
  booktitle={Proceedings of the 30th ACM SIGKDD Conference on Knowledge Discovery and Data Mining},
  pages={199--210},
  year={2024}
}

@inproceedings{jimenezswe,
  title={SWE-bench: Can Language Models Resolve Real-world Github Issues?},
  author={Jimenez, Carlos E and Yang, John and Wettig, Alexander and Yao, Shunyu and Pei, Kexin and Press, Ofir and Narasimhan, Karthik R},
  booktitle={The Twelfth International Conference on Learning Representations}
}

@inproceedings{zhang2024language,
  title={How language model hallucinations can snowball},
  author={Zhang, Muru and Press, Ofir and Merrill, William and Liu, Alisa and Smith, Noah A},
  booktitle={Proceedings of the 41st International Conference on Machine Learning},
  pages={59670--59684},
  year={2024}
}

@inproceedings{aichberger_rethinking_2026,
 author = {Aichberger, Lukas and Schweighofer, Kajetan and Hochreiter, Sepp},
 booktitle = {International Conference on Learning Representations (ICLR)},
 title = {Rethinking {Uncertainty} {Estimation} in {LLMs}: {A} {Principled} {Single}-{Sequence} {Measure}},
 year = {2026}
}

@inproceedings{bang_hallulens_2025,
 author = {Bang, Yejin and Ji, Ziwei and Schelten, Alan and Hartshorn, Anthony and Fowler, Tara and Zhang, Cheng and Cancedda, Nicola and Fung, Pascale},
 booktitle = {Annual Meeting of the Association for Computational Linguistics (ACL)},
 month = {July},
 title = {{HalluLens}: {LLM} {Hallucination} {Benchmark}},
 year = {2025}
}

@inproceedings{chen_inside_2024,
 author = {Chen, Chao and Liu, Kai and Chen, Ze and Gu, Yi and Wu, Yue and Tao, Mingyuan and Fu, Zhihang and Ye, Jieping},
 booktitle = {International Conference on Learning Representations (ICLR)},
 title = {{INSIDE}: {LLMs}' {Internal} {States} {Retain} the {Power} of {Hallucination} {Detection}},
 year = {2024}
}

@inproceedings{cheng_integrative_2025,
 author = {Cheng, Yi and Liang, Xiao and Gong, Yeyun and Xiao, Wen and Wang, Song and Zhang, Yuji and Hou, Wenjun and Xu, Kaishuai and Liu, Wenge and Li, Wenjie and Jiao, Jian and Chen, Qi and Cheng, Peng and Xiong, Wayne},
 booktitle = {International Conference on Learning Representations (ICLR)},
 title = {Integrative {Decoding}: {Improving} {Factuality} via {Implicit} {Self}-consistency},
 year = {2025}
}

@inproceedings{cheng_think_2025,
 author = {Cheng, Xiaoxue and Li, Junyi and Zhao, Xin and Wen, Ji-Rong},
 booktitle = {Findings of the Association for Computational Linguistics: ACL 2025},
 month = {July},
 title = {Think {More}, {Hallucinate} {Less}: {Mitigating} {Hallucinations} via {Dual} {Process} of {Fast} and {Slow} {Thinking}},
 year = {2025}
}

@inproceedings{chuang_dola_2024,
 author = {Chuang, Yung-Sung and Xie, Yujia and Luo, Hongyin and Kim, Yoon and Glass, James R. and He, Pengcheng},
 booktitle = {International Conference on Learning Representations (ICLR)},
 title = {{DoLa}: {Decoding} by {Contrasting} {Layers} {Improves} {Factuality} in {Large} {Language} {Models}},
 year = {2024}
}

@misc{deepseek-ai_deepseek-r1_2025,
 author = {DeepSeek-AI and Guo, Daya and Yang, Dejian and Zhang, Haowei and Song, Junxiao and Zhang, Ruoyu and Xu, Runxin and Zhu, Qihao and Ma, Shirong and Wang, Peiyi and Bi, Xiao and Zhang, Xiaokang and Yu, Xingkai and Wu, Yu and Wu, Z. F. and Gou, Zhibin and Shao, Zhihong and Li, Zhuoshu and Gao, Ziyi and Liu, Aixin and Xue, Bing and Wang, Bingxuan and Wu, Bochao and Feng, Bei and Lu, Chengda and Zhao, Chenggang and Deng, Chengqi and Zhang, Chenyu and Ruan, Chong and Dai, Damai and Chen, Deli and Ji, Dongjie and Li, Erhang and Lin, Fangyun and Dai, Fucong and Luo, Fuli and Hao, Guangbo and Chen, Guanting and Li, Guowei and Zhang, H. and Bao, Han and Xu, Hanwei and Wang, Haocheng and Ding, Honghui and Xin, Huajian and Gao, Huazuo and Qu, Hui and Li, Hui and Guo, Jianzhong and Li, Jiashi and Wang, Jiawei and Chen, Jingchang and Yuan, Jingyang and Qiu, Junjie and Li, Junlong and Cai, J. L. and Ni, Jiaqi and Liang, Jian and Chen, Jin and Dong, Kai and Hu, Kai and Gao, Kaige and Guan, Kang and Huang, Kexin and Yu, Kuai and Wang, Lean and Zhang, Lecong and Zhao, Liang and Wang, Litong and Zhang, Liyue and Xu, Lei and Xia, Leyi and Zhang, Mingchuan and Zhang, Minghua and Tang, Minghui and Li, Meng and Wang, Miaojun and Li, Mingming and Tian, Ning and Huang, Panpan and Zhang, Peng and Wang, Qiancheng and Chen, Qinyu and Du, Qiushi and Ge, Ruiqi and Zhang, Ruisong and Pan, Ruizhe and Wang, Runji and Chen, R. J. and Jin, R. L. and Chen, Ruyi and Lu, Shanghao and Zhou, Shangyan and Chen, Shanhuang and Ye, Shengfeng and Wang, Shiyu and Yu, Shuiping and Zhou, Shunfeng and Pan, Shuting and Li, S. S. and Zhou, Shuang and Wu, Shaoqing and Ye, Shengfeng and Yun, Tao and Pei, Tian and Sun, Tianyu and Wang, T. and Zeng, Wangding and Zhao, Wanjia and Liu, Wen and Liang, Wenfeng and Gao, Wenjun and Yu, Wenqin and Zhang, Wentao and Xiao, W. L. and An, Wei and Liu, Xiaodong and Wang, Xiaohan and Chen, Xiaokang and Nie, Xiaotao and Cheng, Xin and Liu, Xin and Xie, Xin and Liu, Xingchao and Yang, Xinyu and Li, Xinyuan and Su, Xuecheng and Lin, Xuheng and Li, X. Q. and Jin, Xiangyue and Shen, Xiaojin and Chen, Xiaosha and Sun, Xiaowen and Wang, Xiaoxiang and Song, Xinnan and Zhou, Xinyi and Wang, Xianzu and Shan, Xinxia and Li, Y. K. and Wang, Y. Q. and Wei, Y. X. and Zhang, Yang and Xu, Yanhong and Li, Yao and Zhao, Yao and Sun, Yaofeng and Wang, Yaohui and Yu, Yi and Zhang, Yichao and Shi, Yifan and Xiong, Yiliang and He, Ying and Piao, Yishi and Wang, Yisong and Tan, Yixuan and Ma, Yiyang and Liu, Yiyuan and Guo, Yongqiang and Ou, Yuan and Wang, Yuduan and Gong, Yue and Zou, Yuheng and He, Yujia and Xiong, Yunfan and Luo, Yuxiang and You, Yuxiang and Liu, Yuxuan and Zhou, Yuyang and Zhu, Y. X. and Xu, Yanhong and Huang, Yanping and Li, Yaohui and Zheng, Yi and Zhu, Yuchen and Ma, Yunxian and Tang, Ying and Zha, Yukun and Yan, Yuting and Ren, Z. Z. and Ren, Zehui and Sha, Zhangli and Fu, Zhe and Xu, Zhean and Xie, Zhenda and Zhang, Zhengyan and Hao, Zhewen and Ma, Zhicheng and Yan, Zhigang and Wu, Zhiyu and Gu, Zihui and Zhu, Zijia and Liu, Zijun and Li, Zilin and Xie, Ziwei and Song, Ziyang and Pan, Zizheng and Huang, Zhen and Xu, Zhipeng and Zhang, Zhongyu and Zhang, Zhen},
 journal = {arXiv},
 month = {January},
 title = {{DeepSeek}-{R1}: {Incentivizing} {Reasoning} {Capability} in {LLMs} via {Reinforcement} {Learning}},
 year = {2025}
}

@inproceedings{duan_shifting_2024,
 author = {Duan, Jinhao and Cheng, Hao and Wang, Shiqi and Zavalny, Alex and Wang, Chenan and Xu, Renjing and Kailkhura, Bhavya and Xu, Kaidi},
 booktitle = {Annual Meeting of the Association for Computational Linguistics (ACL)},
 title = {Shifting {Attention} to {Relevance}: {Towards} the {Predictive} {Uncertainty} {Quantification} of {Free}-{Form} {Large} {Language} {Models}},
 year = {2024}
}

@article{farquhar_detecting_2024,
 author = {Farquhar, Sebastian and Kossen, Jannik and Kuhn, Lorenz and Gal, Yarin},
 journal = {Nature},
 month = {June},
 title = {Detecting hallucinations in large language models using semantic entropy},
 year = {2024}
}

@inproceedings{gu_mask-dpo_2025,
 author = {Gu, Yuzhe and Zhang, Wenwei and Lyu, Chengqi and Lin, Dahua and Chen, Kai},
 booktitle = {International Conference on Learning Representations (ICLR)},
 title = {Mask-{DPO}: {Generalizable} {Fine}-grained {Factuality} {Alignment} of {LLMs}},
 year = {2025}
}

@inproceedings{he_deberta_2021,
 author = {He, Pengcheng and Liu, Xiaodong and Gao, Jianfeng and Chen, Weizhu},
 booktitle = {International Conference on Learning Representations (ICLR)},
 month = {October},
 title = {{DeBERTa}: {Decoding}-enhanced {BERT} with {Disentangled} {Attention}},
 year = {2021}
}

@inproceedings{huang_factalign_2024,
 author = {Huang, Chao-Wei and Chen, Yun-Nung},
 booktitle = {Findings of the Association for Computational Linguistics: EMNLP 2024},
 month = {November},
 title = {{FactAlign}: {Long}-form {Factuality} {Alignment} of {Large} {Language} {Models}},
 year = {2024}
}

@article{ji_survey_2023,
 author = {Ji, Ziwei and Lee, Nayeon and Frieske, Rita and Yu, Tiezheng and Su, Dan and Xu, Yan and Ishii, Etsuko and Bang, Ye Jin and Madotto, Andrea and Fung, Pascale},
 journal = {ACM Comput. Surv.},
 month = {March},
 title = {Survey of {Hallucination} in {Natural} {Language} {Generation}},
 year = {2023}
}

@misc{kossen_semantic_2024,
 author = {Kossen, Jannik and Han, Jiatong and Razzak, Muhammed and Schut, Lisa and Malik, Shreshth and Gal, Yarin},
 journal = {arXiv},
 month = {June},
 title = {Semantic {Entropy} {Probes}: {Robust} and {Cheap} {Hallucination} {Detection} in {LLMs}},
 year = {2024}
}

@article{lu_towards_2026,
 author = {Lu, Chris and Lu, Cong and Lange, Robert Tjarko and Yamada, Yutaro and Hu, Shengran and Foerster, Jakob and Ha, David and Clune, Jeff},
 journal = {Nature},
 month = {March},
 title = {Towards end-to-end automation of {AI} research},
 year = {2026}
}

@inproceedings{manakul_selfcheckgpt_2023,
 author = {Manakul, Potsawee and Liusie, Adian and Gales, Mark},
 booktitle = {Empirical Methods in Natural Language Processing (EMNLP)},
 title = {{SelfCheckGPT}: {Zero}-{Resource} {Black}-{Box} {Hallucination} {Detection} for {Generative} {Large} {Language} {Models}},
 year = {2023}
}

@inproceedings{min_factscore_2023,
 author = {Min, Sewon and Krishna, Kalpesh and Lyu, Xinxi and Lewis, Mike and Yih, Wen-tau and Koh, Pang and Iyyer, Mohit and Zettlemoyer, Luke and Hajishirzi, Hannaneh},
 booktitle = {Empirical Methods in Natural Language Processing (EMNLP)},
 month = {December},
 title = {{FActScore}: {Fine}-grained {Atomic} {Evaluation} of {Factual} {Precision} in {Long} {Form} {Text} {Generation}},
 year = {2023}
}

@inproceedings{muennighoff_s1_2025,
 author = {Muennighoff, Niklas and Yang, Zitong and Shi, Weijia and Li, Xiang Lisa and Fei-Fei, Li and Hajishirzi, Hannaneh and Zettlemoyer, Luke and Liang, Percy and Candes, Emmanuel and Hashimoto, Tatsunori},
 booktitle = {Empirical Methods in Natural Language Processing (EMNLP)},
 month = {November},
 title = {s1: {Simple} test-time scaling},
 year = {2025}
}

@inproceedings{mundler_self-contradictory_2024,
 author = {Mündler, Niels and He, Jingxuan and Jenko, Slobodan and Vechev, Martin},
 booktitle = {International Conference on Learning Representations (ICLR)},
 title = {Self-contradictory {Hallucinations} of {Large} {Language} {Models}: {Evaluation}, {Detection} and {Mitigation}},
 year = {2024}
}

@misc{obeso_real-time_2025,
 author = {Obeso, Oscar and Arditi, Andy and Ferrando, Javier and Freeman, Joshua and Holmes, Cameron and Nanda, Neel},
 journal = {arXiv},
 month = {August},
 title = {Real-{Time} {Detection} of {Hallucinated} {Entities} in {Long}-{Form} {Generation}},
 year = {2025}
}

@misc{openai_openai_2025,
 author = {OpenAI},
 journal = {arXiv},
 title = {{OpenAI} o3 and o4-mini {System} {Card}},
 year = {2025}
}

@inproceedings{tian_fine-tuning_2024,
 author = {Tian, Katherine and Mitchell, Eric and Yao, Huaxiu and Manning, Christopher D. and Finn, Chelsea},
 booktitle = {International Conference on Learning Representations (ICLR)},
 title = {Fine-{Tuning} {Language} {Models} for {Factuality}},
 year = {2024}
}

@inproceedings{wang_improving_2024,
 author = {Wang, Ante and Song, Linfeng and Peng, Baolin and Jin, Lifeng and Tian, Ye and Mi, Haitao and Su, Jinsong and Yu, Dong},
 booktitle = {Findings of the Association for Computational Linguistics: ACL 2024},
 month = {August},
 title = {Improving {LLM} {Generations} via {Fine}-{Grained} {Self}-{Endorsement}},
 year = {2024}
}

@inproceedings{wei_chain--thought_2022,
 author = {Wei, Jason and Wang, Xuezhi and Schuurmans, Dale and Bosma, Maarten and Ichter, Brian and Xia, Fei and Chi, Ed and Le, Quoc and Zhou, Denny},
 booktitle = {Neural Information Processing Systems (NeurIPS)},
 title = {Chain-of-{Thought} {Prompting} {Elicits} {Reasoning} in {Large} {Language} {Models}},
 year = {2022}
}

@inproceedings{wei_long-form_2024,
 author = {Wei, Jerry and Yang, Chengrun and Song, Xinying and Lu, Yifeng and Hu, Nathan and Huang, Jie and Tran, Dustin and Peng, Daiyi and Liu, Ruibo and Huang, Da and Du, Cosmo and Le, Quoc V.},
 booktitle = {Neural Information Processing Systems (NeurIPS)},
 month = {December},
 title = {Long-form factuality in large language models},
 year = {2024}
}

@misc{yang_hallucinate_2025,
 author = {Yang, Joonho and Yoon, Seunghyun and Chang, Hwan and Kim, Byeongjeong and Lee, Hwanhee},
 journal = {arXiv},
 month = {May},
 title = {Hallucinate at the {Last} in {Long} {Response} {Generation}: {A} {Case} {Study} on {Long} {Document} {Summarization}},
 year = {2025}
}

@misc{yang_qwen3_2025,
 author = {Yang, An and Li, Anfeng and Yang, Baosong and Zhang, Beichen and Hui, Binyuan and Zheng, Bo and Yu, Bowen and Gao, Chang and Huang, Chengen and Lv, Chenxu and Zheng, Chujie and Liu, Dayiheng and Zhou, Fan and Huang, Fei and Hu, Feng and Ge, Hao and Wei, Haoran and Lin, Huan and Tang, Jialong and Yang, Jian and Tu, Jianhong and Zhang, Jianwei and Yang, Jianxin and Yang, Jiaxi and Zhou, Jing and Zhou, Jingren and Lin, Junyang and Dang, Kai and Bao, Keqin and Yang, Kexin and Yu, Le and Deng, Lianghao and Li, Mei and Xue, Mingfeng and Li, Mingze and Zhang, Pei and Wang, Peng and Zhu, Qin and Men, Rui and Gao, Ruize and Liu, Shixuan and Luo, Shuang and Li, Tianhao and Tang, Tianyi and Yin, Wenbiao and Ren, Xingzhang and Wang, Xinyu and Zhang, Xinyu and Ren, Xuancheng and Fan, Yang and Su, Yang and Zhang, Yichang and Zhang, Yinger and Wan, Yu and Liu, Yuqiong and Wang, Zekun and Cui, Zeyu and Zhang, Zhenru and Zhou, Zhipeng and Qiu, Zihan},
 journal = {arXiv},
 month = {May},
 title = {Qwen3 {Technical} {Report}},
 year = {2025}
}

@inproceedings{yao_tree_2023,
 author = {Yao, Shunyu and Yu, Dian and Zhao, Jeffrey and Shafran, Izhak and Griffiths, Tom and Cao, Yuan and Narasimhan, Karthik},
 booktitle = {Neural Information Processing Systems (NeurIPS)},
 month = {December},
 title = {Tree of {Thoughts}: {Deliberate} {Problem} {Solving} with {Large} {Language} {Models}},
 year = {2023}
}

@inproceedings{zhao_how_2025,
 author = {Zhao, James Xu and Liu, Jimmy Z. J. and Hooi, Bryan and Ng, See-Kiong},
 booktitle = {ACL},
 month = {May},
 title = {How {Does} {Response} {Length} {Affect} {Long}-{Form} {Factuality}},
 year = {2025}
}
\bibliographystyle{icml2026}

\newpage
\appendix
\onecolumn

\section{Prompts}
\label{sec: prompts}

\begin{figure*}[t]
    \centering
    \includegraphics[width=0.75\linewidth]{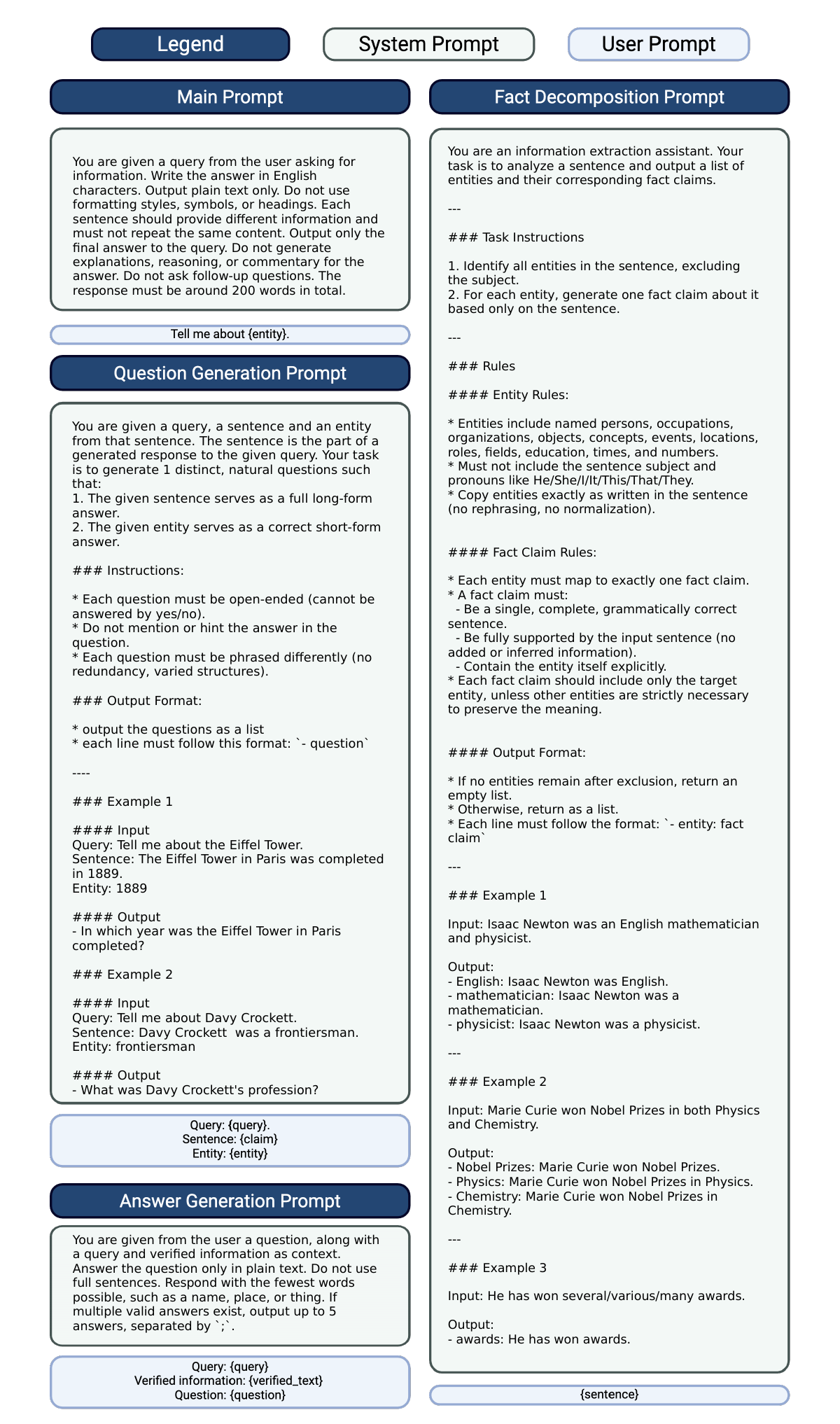}
    \caption{Set of prompts for various parts of the pipeline.}
    \label{fig:prompts1}
\end{figure*}

\begin{figure*}[t]
    \centering
    \includegraphics[width=0.7\linewidth]{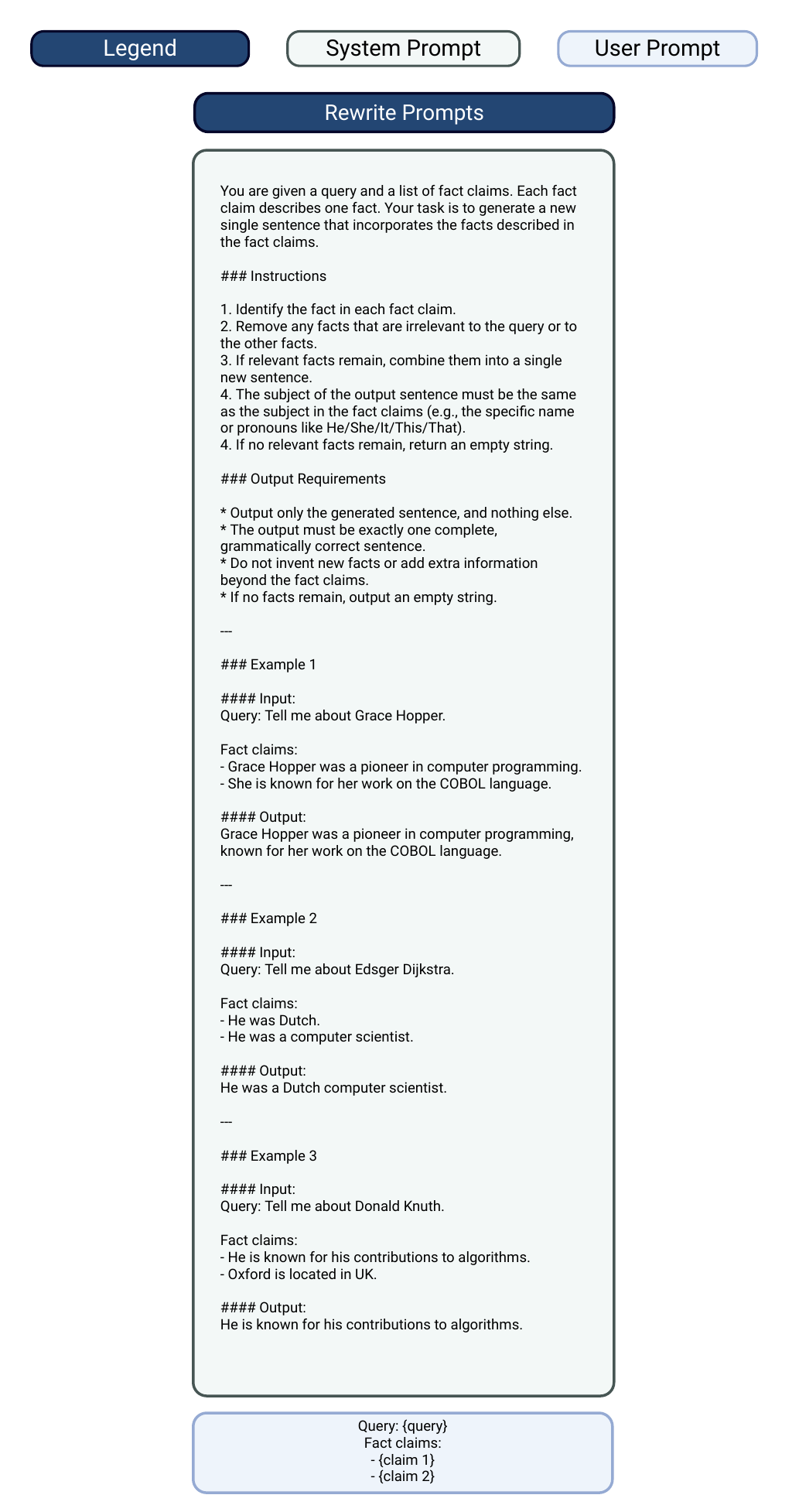}
    \caption{Additional prompt if rewrite is enabled.}
    \label{fig:prompts2}
\end{figure*}

Prompts are given in \cref{fig:prompts1} and \cref{fig:prompts2}.

\section{Experiments}

\subsection{Configuration}
\label{app: configuration}

FactScore \citep{min_factscore_2023} evaluates LLMs by generating biographies for 182 individuals \footnote{We exclude one individual named Focus... from the original dataset due to confusion with the band of the same name Focus and complications caused by special punctuation.} (the labeled split), spanning diverse demographics and varying levels of rarity. 
Each generation is decomposed into atomic facts, which are verified against a reliable knowledge source, in this case a pre-saved Wikipedia. GPT-5 was used as the backend LLM for the benchmark. The prompts for query are given in \cref{sec: prompts}. 
The results are reported from a single run due to the high API cost of benchmark evaluation.

LongFact \citep{wei_long-form_2024} is a benchmark for long-form factuality with two key differences from FactScore. First, it includes thousands of questions across 38 topics, though we used only a subset of 140 prompts following \citet{zhao_how_2025} due to resource constraints (evaluating a single generation costs at least \$0.19 \citep{wei_long-form_2024}). Second, it relies on results from online Google Search rather than a pre-saved Wikipedia as the knowledge source. GPT-3.5-turbo-0125 is used as the backend LLM. The experimental setup follows that of FactScore in \cref{sec: results hallucination mitigation}. 

We describe here the hyperparameters of our method. The maximum number of tolerated consecutive hallucinated sentences sampling, $N$, is 10 across all setups. In \cref{tab: factscore}, the number of probe questions, $Q$, the number of answers, $A$, and the semantic entropy threshold, $\theta$, are 1, 3, 0.7 for Ours-Resp with Llama3.1-8B-Instruct; 3, 3, 0.3 for Ours-Info with Llama3.1-8B-Instruct; 2, 3, 0.3 for Ours-Prec with Llama3.1-8B-Instruct; 1, 5, 0.7 for Ours-Resp with Qwen3-32B; 2, 3, 0.2 for Ours-Info with Qwen3-32B; 2, 7, 0.6 for Ours-Prec with Qwen3-32B. In \cref{tab: factscore 200words}, the number of probe questions, $Q$, the number of answers, $A$, and the semantic entropy threshold, $\theta$, are 1, 3, 0.5 for Ours-Info; 2, 3, 0.3 for Ours-Prec. In \cref{tab: longfact}, the number of probe questions, $Q$, the number of answers, $A$, and the semantic entropy threshold, $\theta$, are 3, 3, 0.3 for without length constraint; 2, 3, 0.5 for 200-words constraint. In \cref{tab: ablation study}, the number of probe questions, $Q$, the number of answers, $A$, and the semantic entropy threshold, $\theta$, are 1, 3, 0.5 for all. All above hyperparameters, except for the ones for LongFact, are found through a coarse grid search. 

Meanwhile, our method can deliver substantial performance improvements across diverse models and datasets using fixed hyperparameter values, and these gains can be further enhanced by tuning the hyperparameters for specific models or data. For example, we find that Q=3, A=3, and theta=0.3 consistently achieve the highest or near-highest factual accuracy across all settings, while Q=1, A=3, and theta=0.5 provide strong accuracy gains with moderate computational cost.

\subsection{NLI Model in Semantic Entropy}
\label{app: nli model}
The NLI model we used is DeBERTa V2 \citep{he_deberta_2021} with 900M parameters, identical to the one used in Semantic Entropy. Specifically, we used the deberta-v2-xlarge-mnli checkpoint from HuggingFace. It was trained on the following datasets: Wikipedia (English dump; 12GB), BookCorpus (Zhu et al., 2015; 6GB), OPENWEBTEXT (public Reddit content, Gokaslan \& Cohen, 2019; 38GB), and STORIES (a CommonCrawl subset, Trinh \& Le, 2018; 31GB). These sources span diverse domains, making the model effectively domain-agnostic.

\section{Additional Results}

\subsection{Standard deviation}
\label{app: standard deviation}

To assess robustness, we repeat the experiments for Qwen3-32B and Llama3.1-8B in \cref{tab: factscore} three times and measure the mean and standard deviation in \cref{tab: std}, which confirm the robustness and consistency of our results.

We originally reported single-run results because the performance margins over baselines are large and stable across models, and the evaluation is expensive due to API costs. This is also consistent with common practice in prior work under similar constraints.

\begin{table}[h]
\centering
\caption{Standard deviation of performance on the FactScore benchmark without constraints on response length.}
\label{tab: std}
\begin{tabular}{@{}lcccc@{}}
\toprule
\multicolumn{1}{c}{Method} & Resp. (\%)     & Unsup.       & Supp.         & FPrec. (\%)   \\ \midrule
\multicolumn{5}{c}{Llama3.1-8B}                                                           \\ \midrule
Ours-Resp                  & \textbf{99.27} $\pm$ 0.32 & 3.31 $\pm$ 0.10       & 5.60 $\pm$ 0.30         & 62.82 $\pm$ 1.55        \\
Ours-Info                  & 88.46 $\pm$ 0.00        & 1.82 $\pm$ 0.13        & 5.46 $\pm$ 0.41         & 75.01 $\pm$ 1.52        \\
Ours-Prec                  & 77.02 $\pm$ 1.22        & \textbf{1.46} $\pm$ 0.07 & 4.91 $\pm$ 0.12         & \textbf{77.02} $\pm$ 1.22 \\ \midrule
\multicolumn{5}{c}{Qwen3-32B}                                                             \\ \midrule
Ours-Resp                  & 97.25 $\pm$ 0.95         & 5.55 $\pm$ 0.15         & 10.74 $\pm$ 0.62        & 65.90 $\pm$ 1.28    \\
Ours-Info                  & 91.58 $\pm$ 1.14        & 4.31 $\pm$ 0.22        & \textbf{11.33} $\pm$ 0.33 & 72.43 $\pm$ 1.16        \\
Ours-Prec                  & 82.97 $\pm$ 1.98        & \textbf{3.20} $\pm$ 0.26 & 10.89 $\pm$ 0.24        & \textbf{77.32} $\pm$ 1.36 \\ \bottomrule
\end{tabular}
\end{table}

\section{Discussion}

\subsection{How does SHARS compare with retrieval-based methods?}

SHARS and retrieval-based approaches (e.g., RAG) address complementary aspects of hallucination: RAG improves factuality by expanding the model’s knowledge, while SHARS improves how the model utilizes its parametric knowledge. Moreover, long-form factuality benchmarks often rely on external knowledge sources as ground-truth (e.g., FactScore uses Wikipedia), making direct comparisons between these approaches less meaningful.

Importantly, the two can be combined in multiple ways. First, SHARS can be used in an adaptive pipeline, where retrieval is triggered only when repeated rejection or abstention indicates insufficient knowledge, improving efficiency and reducing unnecessary retrieval noise. Second, SHARS can be applied on top of retrieval-augmented generation to filter or correct hallucinations caused by imperfect retrieval or intrinsic model errors.

\end{document}